\documentclass{article}

\PassOptionsToPackage{numbers,sort&compress}{natbib}
\usepackage[main, final]{neurips_2026}

\usepackage[utf8]{inputenc} %
\usepackage[T1]{fontenc}    %
\usepackage{hyperref}       %
\usepackage{url}            %
\usepackage{booktabs}       %
\usepackage{amsfonts}       %
\usepackage{nicefrac}       %
\usepackage{microtype}      %
\usepackage{amsmath}
\usepackage{graphicx}
\usepackage{multirow}
\usepackage[table]{xcolor}

\newif\ifdraft

\ifdraft
 \newcommand{\PF}[1]{{\color{red}{\bf PF: #1}}}
 
  \newcommand{\AF}[1]{{\color{blue}{\bf AF: #1}}} 
 \newcommand{\af}[1]{{\color{blue} #1}}
 \newcommand{\HL}[1]{{\color{orange}{\bf HL: #1}}} 

\newcommand{\CD}[1]{{\color{purple}{\bf CD: #1}}} 

\newcommand{\NT}[1]{{\color{teal}{\bf NT: #1}}} 

\newcommand{\MX}[1]{{\color{cyan}{\bf MX: #1}}} 

\else
 \newcommand{\PF}[1]{}
 
 \newcommand{\AF}[1]{} 
 \newcommand{\af}[1]{#1}
 \newcommand{\HL}[1]{}
 
 \newcommand{\CD}[1]{}
 
 \newcommand{\NT}[1]{}
 
 \newcommand{\MX}[1]{}
 
\fi

\newcommand{\first}{\cellcolor{red!25}}
\newcommand{\second}{\cellcolor{orange!25}}
\newcommand{\third}{\cellcolor{yellow!35}}

\title{Revisiting Multi-View Stereo: A Sequence-to-Sequence Formulation}

\author{
Aoxiang Fan \quad Corentin Dumery \quad Nicolas Talabot \quad Pascal Fua\\
CVLab, EPFL\\
Switzerland\\
{\texttt{\{aoxiang.fan, corentin.dumery, nicolas.talabot, pascal.fua\}@epfl.ch}}
}

\begin{document}

\maketitle

\begin{abstract}
  Computing accurate geometry from multi-view images is a fundamental problem in computer vision. Recent feed-forward (FF) models jointly estimate 3D geometry and camera parameters, but they typically suffer from geometry distortion caused by reconstruction ambiguity, even when ground-truth camera parameters are supplied. In this paper, we study the multi-view stereo (MVS) problem with known camera parameters and propose a novel approach that bridges conventional MVS and FF methods. Rather than casting MVS as a sequence-to-one mapping that predicts depth only for a single reference view, we reformulate it as a sequence-to-sequence task, akin to FF models, that jointly predicts geometry for all input views. We introduce a global transformer-based architecture with two components that explicitly exploit camera-induced priors: ray-map embeddings that inject camera parameters into image patch tokens, making the transformer camera-aware, and a unified global cost volume that replaces conventional per-view cost volumes to jointly capture 3D structure across all views. Extensive experiments on multiple public benchmarks show our approach achieves state-of-the-art performance, surpassing both MVS and FF reconstruction baselines\af{.}
\end{abstract}
\section{Introduction}

3D reconstruction from multi-view images is a fundamental and long-standing problem in computer vision that has seen significant progress in recent years. Conventionally, the reconstruction task is addressed by Structure-from-Motion (SfM) and Multi-view Stereo (MVS), where both classical and learning-based methods have been developed. Classical methods~\citep{pix4d,schonberger2016structure,moulon2016openmvg,galliani2015massively,tola2012efficient,furukawa2009accurate} are able to reconstruct very accurate 3D points by extracting image correspondences and 3D triangulation. However, this often results in sparse and incomplete reconstructions, requiring complex densification techniques in post-processing~\citep{galliani2015massively,tola2012efficient,furukawa2009accurate}.

Learning-based MVS methods~\citep{yao2018mvsnet,ding2022transmvsnet,cao2024mvsformer++,izquierdo2025mvsanywhere} have been proposed to address this issue, leveraging deep learning to infer per-pixel dense depth maps directly from multi-view images. The key innovation of these methods is \textbf{cost volumes} as a camera-induced prior, which are constructed by regularly sampling 3D points along per-pixel depth dimension and store view-consistency information for each point. In this way, cost volumes offer very effective regularization for geometry estimation. Nevertheless, they are typically constructed on a per-view basis, effectively reducing the problem to a sequence-to-one formulation that predicts only the depth map of a reference view, even when multiple views are provided as input. The sequence-to-one design not only reduces efficiency but also restricts geometric estimation accuracy due to the lack of global alignment modeling.

Among the most notable advances in recent years are feed-forward (FF) models~\citep{wang2025vggt,yang2025fast3r,wang2025pi,keetha2025mapanything,lin2025depth,wang2026vggt,burzio2026d} designed for general settings and adopt a sequence-to-sequence formulation to jointly estimate camera parameters and depth/point maps for all input views. However, due to the inherent difficulty of this joint estimation problem, the resulting reconstructions are often less accurate or more distorted. This remains true even when ground-truth camera parameters are provided, as the model still attempts to estimate and update them.

\begin{figure}[t]
    \centering
    \includegraphics[width=\linewidth]{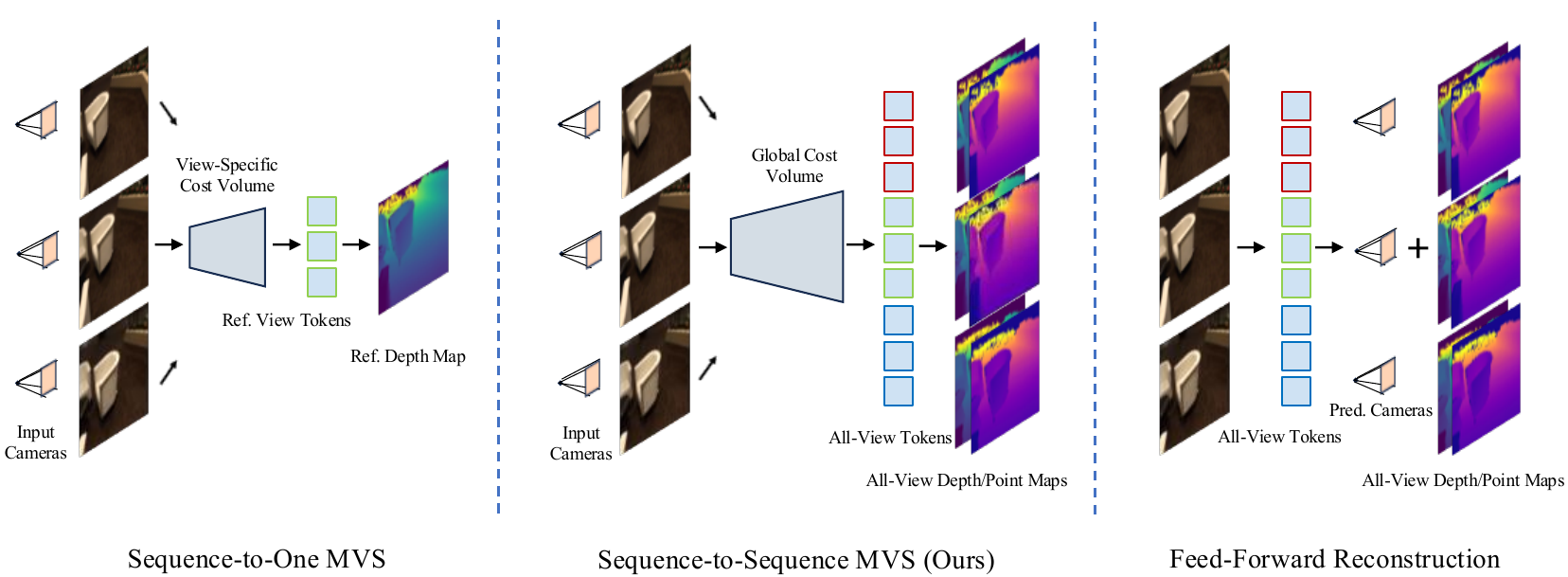}
    \caption{The differences of formulation between sequence-to-one multi-view stereo methods, our proposed sequence-to-sequence multi-view stereo method and feed-forward reconstruction methods.} 
    \label{fig:s2s}
\end{figure}

In this paper, we study the MVS problem that conditions on known camera parameters. These parameters provide greater control over the reconstruction process and can be readily obtained in practice using SfM or other camera calibration methods. Our approach bridges MVS and FF methods: like feed-forward models, it adopts a sequence-to-sequence formulation with a global Transformer-based architecture, while incorporating camera-induced priors to ensure geometric accuracy. Specifically, we consider two types of such priors. Firstly and naturally, we need to incorporate camera parameters into the transformer-based architecture to make the model camera-aware. Secondly, we propose to adapt the conventional cost-volume idea to our sequence-to-sequence setting, providing valuable regularization over the underlying 3D scene geometry.  We highlight the differences of our formulation and the existing ones in Fig.~\ref{fig:s2s}.

\begin{figure}[t]
    \centering
    \includegraphics[width=\linewidth]{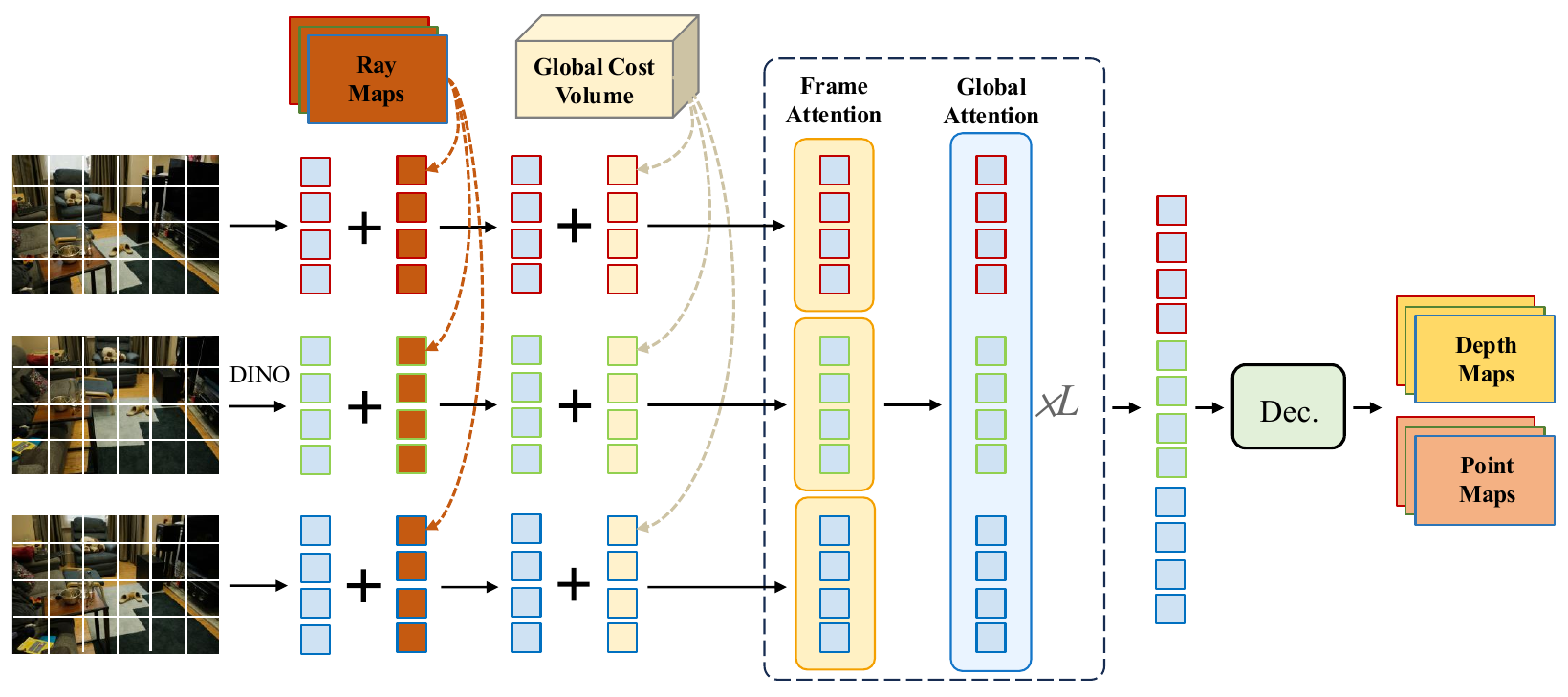}
    \caption{Overview of the proposed sequence-to-sequence multi-view stereo method. For each input view, we extract view tokens using DINOv2 backbone, and incorporate camera-induced priors into the model with ray-map embeddings and global-cost-volume embeddings. The tokens are then fed into recurring attention layers to aggregate frame and global information, which are fed to Decoding heads to produce depth maps and point maps of each input view. } 
    \label{fig:main_method}
\end{figure}

To integrate these priors into our framework, we introduce two key designs. First, inspired by~\citep{lin2025depth}, we transform the camera intrinsics and extrinsics into ray maps, which are then tokenized and merged with image tokens to encode camera information. Second, we introduce a Global Cost Volume, which, unlike conventional cost volumes that have to be repeatedly constructed per-view, captures the global 3D structure of the scene across all views in a unified representation. We provide a diagram illustration of our method in Fig.~\ref{fig:main_method}.

Our model operates in a sequence-to-sequence manner, in contrast to existing sequence-to-one MVS methods that also incorporate camera-induced priors. Experimental results demonstrate that it achieves superior reconstruction results compared to both MVS approaches and recent sequence-to-sequence feed-forward models, establishing a new state of the art for multi-view reconstruction with known camera parameters.
\section{Related Works}

\textbf{Structure-from-Motion.} Structure-from-Motion (SfM)~\citep{hartley2003multiple,ozyecsil2017survey,schonberger2016structure,pan2024global,moulon2016openmvg,pix4d,moulon2016openmvg} aims to jointly estimate camera parameters and sparse 3D structure from multi-view images. Classical pipelines, such as COLMAP~\citep{schonberger2016structure}, Pix4D~\citep{pix4d} and OpenMVG~\citep{moulon2016openmvg}, decompose the problem into stages including feature extraction, matching, triangulation, and bundle adjustment, and remain strong baselines under well-constrained settings~\citep{lowe1999object,bay2008speeded,rublee2011orb,hartley2003multiple}. Recent progress has focused on improving robustness through learned components, particularly in feature detection and matching. These include sparse matching pipelines with local feature detection~\citep{detone2018superpoint,tyszkiewicz2020disk} and matching~\citep{yi2018learning,sarlin2020superglue,lindenberger2023lightglue}, and dense matching pipelines without feature detectors~\citep{sun2021loftr,truong2020glu,edstedt2023dkm,edstedt2024roma}. Beyond modular pipelines, end-to-end differentiable approaches such as VGGSfM~\citep{wang2024vggsfm} jointly optimize camera poses and scene structure, improving performance in challenging scenarios. Recently, VidMap~\citep{pataki2026vidmap} has been proposed for video-based SfM by leveraging temporal information. Nevertheless, dense reconstruction remains a challenge in this line of work, subject to expensive and complex post-processing steps~\citep{galliani2015massively,openmvs,tola2012efficient,furukawa2009accurate}.

\textbf{Learning-based Multi-View Stereo.} Recent MVS methods are mostly learning-based~\citep{yao2018mvsnet,wang2021patchmatchnet,gu2020cascade,ma2022multiview,peng2022rethinking,ding2022transmvsnet,cao2024mvsformer++,izquierdo2025mvsanywhere}, focusing on dense reconstruction given calibrated cameras typically provided by SfM. Traditional methods rely on handcrafted priors or global optimization, while learning-based approaches, such as MVSNet~\citep{yao2018mvsnet}, leverage cost volumes and 3D convolutional networks to infer accurate depth maps. These models significantly improve reconstruction quality, particularly in challenging regions like texure-less structures or wide baselines. Recent work, including MVSA~\citep{izquierdo2025mvsanywhere}, emphasizes generalization across diverse scenes and viewing conditions. Despite these advances, most MVS methods construct cost volumes on a per-view basis, limiting their ability to capture global scene structure. Moreover, their sequence-to-one formulation restricts predictions to a reference view, leaving multi-view consistency to be handled implicitly.

\textbf{Feed-Forward Reconstruction.} Feed-Forward approaches reformulate multi-view reconstruction as a direct prediction problem using neural networks, often built on transformer architectures. Early works focus on pairwise inputs, with methods such as DUSt3R~\citep{wang2024dust3r} directly predicting aligned point maps without requiring camera parameters, and MASt3R~\citep{leroy2024grounding} improving geometric consistency and scale estimation. Multi-view feed-forward models, including VGGT~\citep{wang2025vggt}, Fast3R~\citep{yang2025fast3r}, MapAnything~\citep{keetha2025mapanything}, $\pi^3$~\citep{wang2025pi}, DepthAnything3~\citep{lin2025depth}, VGGT-$\Omega$~\citep{wang2026vggt}, and DVLT~\citep{burzio2026d}, aggregate features across views using global attention to jointly estimate camera parameters and scene geometry in a single forward pass. However, global attention incurs quadratic complexity with respect to the number of input views, limiting scalability. Approaches such as FastVGGT~\citep{shen2025fastvggt} and SparseVGGT~\citep{wang2025faster} improve efficiency through token reduction or sparse attention. Despite these advances, jointly estimating camera parameters and geometry remains highly challenging, and such models are usually prone to reconstruction inaccuracies or distortion.

\section{Method}

In this section, we describe the proposed method, termed Sequence-to-Sequence MVS (S2S-MVS). We start by introducing the setting and formulating the addressed problem in Sec. \ref{sec:problem_formulation}, as well as explaining the basic architecture of our framework. Then, we separately explain how the two camera-induced priors can be incorporated into our basic architecture in Sec. \ref{sec:camera_parameter_encoding} and Sec. \ref{sec:cost_volume_construction}.

\subsection{Problem Formulation}
\label{sec:problem_formulation}

\textbf{Definitions and Notations.} We consider as input a sequence of $V$ RGB images $(I_i)_{i=1}^V$, where $I_i \in \mathbb{R}^{3\times H\times W}$, 
capturing a 3D scene. Differently from most feed-forward models, we follow MVS setting to assume camera parameters are known, which are typically from SfM models and is the premise for studying camera-induced priors. Formally, our model's output is written as
\begin{equation}
f\left((I_i)_{i=1}^V, (\mathbf{g}_i)_{i=1}^V\right)
=
\left(
D_i,
P_i
\right)_{i=1}^V,
\end{equation}
where $\mathbf{g}_i$ denotes the known camera parameters of image $I_i$.
For each frame $I_i$, the model predicts a depth map $D_i \in \mathbb{R}^{H\times W}$ and a point map $P_i \in \mathbb{R}^{3\times H\times W}$. The known \textbf{camera parameters} $\mathbf{g}_i$ include both extrinsics $(\mathbf{R}_i, \mathbf{t}_i)_{i=1}^{V}$ and intrinsics $(\mathbf{K}_i)_{i=1}^{V}$. 

We define the image domain of $I_i$ as $\mathcal{I}(I_i)=\{1,\dots,H\}\times\{1,\dots,W\}$, namely, the set of all pixel coordinates. The \textbf{depth map} $D_i$ assigns to each pixel $\mathbf{y} \in \mathcal{I}(I_i)$ a positive scalar depth value $D_i(\mathbf{y})\in\mathbb{R}^+$ with respect to the $i$-th camera. The \textbf{point map} $P_i$ associates each pixel $\mathbf{y}$ with its corresponding 3D location $P_i(\mathbf{y})\in\mathbb{R}^3$. Following VGGT~\citep{wang2025vggt}, these point maps are defined in a \emph{viewpoint-invariant} manner: all 3D points are expressed in the coordinate system of the first camera $\mathbf{g}_1$, which serves as the world reference frame.

For sequence-to-sequence modeling, our network employs a global transformer architecture, following existing feed-forward models. Next, we explain this base architecture.

\textbf{Image Feature Encoding.} This module produces powerful monocular feature representations for \af{input} images. Specifically, we adopt the \af{ViT-L/16 encoder of DINOv3~\citep{simeoni2025dinov3}, initialized with its pretrained weights}. The model processes images using $k \times k$ patches\af{ with $k=16$}, yielding feature maps of spatial size $\frac{H}{k} \times \frac{W}{k}$. As a result, each image is decomposed into a collection of $K$ token embeddings, denoted by $\mathrm{t}_i^I \in \mathbb{R}^{K\times C}$. \af{We take these tokens from four intermediate layers of the encoder, which together with the cost volume, form the input tokens of the alternating-attention network, as explained in Sec.~\ref{sec:cost_volume_construction}.}

\textbf{Frame and Global Attention.} Consistent with existing works~\citep{wang2025vggt,lin2025depth,burzio2026d}, our transformer architecture alternates between frame-wise self-attention and global self-attention. Frame-wise self-attention is applied to the tokens of each input image independently, which focuses on per-view scene information. Global self-attention is applied to all tokens across all input images, aggregating crucial multi-view information to facilitate reconstruction.

\textbf{Dense Predictions.} We take \af{the processed tokens from four attention layers spread across the network, which are fused and} progressively upsampled using learned convolutional filters. \af{Following} VGGT-$\Omega$~\citep{wang2026vggt}, we replace the blocks operating above 1/4 of the input resolution with a lightweight upsampling head via a single MLP followed by a pixel-shuffle operator to output full-resolution dense \af{depth maps $(D_i)_{i=1}^V$, from which the point maps $(P_i)_{i=1}^V$ are obtained with the ray maps (Sec.~\ref{sec:camera_parameter_encoding}). In addition to depth maps,} the same head outputs a per-pixel aleatoric uncertainty map $\Sigma_i \in \mathbb{R}^{H\times W}$ for each view, obtained from an extra output channel. This uncertainty map reflects the model's confidence in its own depth and point predictions and is used to reweight the regression losses, as detailed next.

\subsection{Camera Parameter Encoding}
\label{sec:camera_parameter_encoding}

Camera parameters are typically expressed in the literature in two ways. In VGGT~\citep{wang2025vggt}, camera tokens are defined for each frame and serve as special tokens that participate in both frame and global attentions. Alternatively, in DepthAnything3~\citep{lin2025depth}, a depth-ray representation is used as the regression objective. 

In our setting, we assume the camera parameters are known, which allows us to encode camera information from the beginning. We opt for a ray map embedding approach over \af{regressed} camera tokens, which directly injects information in a per-patch manner and stabilizes training.

For each pixel $\mathbf{y}$, we define its camera ray $\mathbf{r} \in \mathbb{R}^6$ as the concatenation of a ray origin $\mathbf{t} \in \mathbb{R}^3$ and a ray direction $\mathbf{d} \in \mathbb{R}^3$, i.e., $\mathbf{r} = (\mathbf{t}, \mathbf{d})$. The ray direction is computed by first backprojecting $\mathbf{y}$ into the camera coordinate system and then transforming it into the world frame:
\[
\mathbf{d} = \mathbf{R} \mathbf{K}^{-1} \mathbf{y} .
\]
where $\mathbf{R}$ and $\mathbf{t}$ represent the rotation and translation in world frame. By collecting the ray origins and directions over all pixels, we obtain a dense ray map $\mathbf{M} \in \mathbb{R}^{H \times W \times 6}$. Note that $\mathbf{d}$ is left unnormalized, allowing its norm to retain the corresponding projection scale. Given the predicted depth, the associated 3D point in world coordinates can then be recovered as
\[
P = {\mathbf{t}} + D(\mathbf{y}) \cdot \mathbf{d}.
\]
With this representation, point maps can be generated in a unified and efficient manner by combining the predicted depth and ray maps using element-wise operations.

\af{\textbf{Incorporation into Transformer.} We apply an element-wise signed logarithm $\mathrm{sign}(\mathbf{M}) \log(1 + |\mathbf{M}|)$ to the ray map, patchify it with the image patch size and embed each patch linearly into ray tokens $\mathrm{r}^I = \bigcup_{i=1}^{V}\{\mathrm{r}^I_i\}$. These tokens are added to the patch tokens after the frame-wise attention of every layer $\ell$, $\mathrm{t}^{(\ell)} \leftarrow \mathrm{t}^{(\ell)} + \tanh(\gamma_\ell) \, \mathrm{r}^I$, where the learnable scalar $\gamma_\ell$ is initialized to zero so that camera information is introduced gradually.}

\subsection{Cost Volume Construction}
\label{sec:cost_volume_construction}

Cost volume is a pivotal module in most existing learning-based MVS methods, providing crucial camera-induced scene priors for multi-view reconstruction. Next, we will explain the conventional cost volume construction first, then turn to our proposed global cost volume.

\subsubsection{Per-View Cost Volume}
Conventionally, the construction of a cost volume needs a reference view and a set of hypothesized depth values $(d_i)_{i=1}^{T}$ to be predefined~\citep{yao2018mvsnet,ding2022transmvsnet,cao2024mvsformer++,izquierdo2025mvsanywhere}. To capture low-level features from images, a convolutional backbone is usually used for cost volume construction, giving feature maps $(\mathcal{F}_i)_{i=1}^V$. The cost volume is constructed by warping feature maps $\mathcal{F}_i$ from each source view to the reference one using hypothesized depth values $(d_i)_{i=1}^{T}$. 

The warping is formulated as a standard 3D reprojection process. Given a pixel $\mathbf{y}$ in the reference image and a depth hypothesis $d$, its projected location $\hat{\mathbf{y}}$ in the source image is computed. Pair-wise feature correlation at position $\mathbf{y}$ is 
\begin{equation}
    c^{(d)}_i(\mathbf{y}) \; = \;<\mathcal{F}_0(\mathbf{y}),\hat{\mathcal{F}}^{(d)}_i(\mathbf{y})>,
\end{equation}
where $\hat{\mathcal{F}}^{(d)}_i$ denotes the warped $i$-th source feature map at depth $d$. The correlation values of each source view is then aggregated, for instance by taking an average, to a scalar cost volume in shape $T \times \frac{H}{s} \times \frac{W}{s}$, where $s$ is the pixel-wise subsampling ratio. 

\subsubsection{Global Cost Volume}
\af{Repeating the per-view construction with every view as the reference makes the cost grow quadratically with $V$. We instead build a single global cost volume shared by all views.}

\af{\textbf{Cost Value Computation.} \label{sec:cost_value_computation}
We drop the reference view and define a cost value for any 3D point $\mathbf{p}$ in the world frame. Projecting $\mathbf{p}$ into view $i$ gives $\lambda_i \tilde{\mathbf{y}}_i = \mathbf{K}_i(\mathbf{R}_i\mathbf{p}+\mathbf{t}_i)$, where $\tilde{\mathbf{y}}_i$ are the homogeneous coordinates of the projection $\mathbf{y}_i$ and $\lambda_i$ is the depth of $\mathbf{p}$. Let $\mathcal{V}(\mathbf{p})$ be the set of $k$ views with $\lambda_i > 0$ and $\mathbf{y}_i$ inside the image, and $\mathbf{f}_i = \mathcal{F}_i(\mathbf{y}_i)$ the bilinearly sampled features. The cost value is the mean correlation over all pairs of distinct views that see $\mathbf{p}$:
\begin{equation}
    c(\mathbf{p}) \; = \; \frac{1}{k(k-1)} \sum_{\substack{i, j \in \mathcal{V}(\mathbf{p}) \\ i \neq j}} <\mathbf{f}_i, \mathbf{f}_j>,
    \label{eq:global_cost}
\end{equation}
with $c(\mathbf{p}) = 0$ if $k < 2$. It treats all views symmetrically, and since $\sum_{i \neq j} <\mathbf{f}_i, \mathbf{f}_j> = \|\sum_i \mathbf{f}_i\|^2 - \sum_i \|\mathbf{f}_i\|^2$, it can be accumulated one view at a time with memory independent of $V$.}

\af{\textbf{Scene Representation in Contracted Space.} As unbounded scenes offer no bounding box, we place $\mathbf{C}$ on a uniform grid in a contracted space. We back-project $T$ log-spaced depth hypotheses $(d_t)_{t=1}^{T}$ at every pixel of every view, normalize these frustum samples by their median $\mathbf{o}$ and median distance $\rho$ to it, and apply the $L_\infty$ contraction of mip-NeRF~360~\citep{barron2022mip}, which maps $\mathbb{R}^3$ onto $[-2, 2]^3$:
\begin{equation}
    \mathrm{contract}(\mathbf{x}) = \mathbf{x} \;\text{ if } \|\mathbf{x}\|_\infty \leq 1, \quad \left(2 - \tfrac{1}{\|\mathbf{x}\|_\infty}\right) \tfrac{\mathbf{x}}{\|\mathbf{x}\|_\infty} \;\text{ otherwise}, \qquad \mathbf{x} = \tfrac{\mathbf{p} - \mathbf{o}}{\rho}.
\end{equation}
$\mathbf{C}$ is obtained by evaluating Eq.~\eqref{eq:global_cost} on a uniform grid over $[-2, 2]^3$ whose size is independent of $V$, and each view-specific cost volume $\mathbf{C}_i \in \mathbb{R}^{T \times \frac{H}{s} \times \frac{W}{s}}$ is read from it by trilinear interpolation, so the correlation cost grows only linearly with $V$.}

\af{\textbf{Incorporation into a Transformer.} Similarly to MVSA~\citep{izquierdo2025mvsanywhere}, a convolutional network $\Phi$ of residual blocks downsamples $\mathbf{C}_i$, with the $T$ hypotheses as channels, from $\frac{H}{4} \times \frac{W}{4}$ to the token resolution $\frac{H}{k} \times \frac{W}{k}$, fusing the image features of the two shallower encoder layers at $\frac{1}{4}$ and $\frac{1}{8}$ resolution and of the two deeper ones at $\frac{1}{16}$:
\[
\mathrm{t}^I_i = \Phi\left(\mathbf{C}_i, \mathrm{t}_i^{I,(1)}, \dots, \mathrm{t}_i^{I,(4)}\right).
\]
The resulting tokens are fed into the frame and global attention layers and eventually decoded into depth maps and point maps.}

\subsection{Training}
\label{sec:training}

\textbf{Training Loss.}
Our loss includes a depth regression loss term, a point map regression term, and a normal loss term:
\begin{align}
    \mathcal{L} = \mathcal{L}_{\text{depth}} + \mathcal{L}_{\text{pmap}} + \mathcal{L}_{\text{normal}}.
\end{align}
For the depth regression loss, we need to normalize the predicted depths to be consistent with the construction of cost volumes. Specifically, depth is estimated by applying a sigmoid function $\sigma$ to the logits $x_{ij}$ returned by prediction heads, which is then scaled by the depth range defined by the cost-volume depth-sampling bounds:
\begin{align}
\hat{D}_{i,j} = \exp\left(\log(d_{1}) + \log(d_{T} / d_{1}) \cdot \sigma(x_{ij})\right).
\end{align}
Rather than treating every pixel as equally reliable, we let the model down-weight regions it is uncertain about (e.g., due to occlusions or textureless surfaces) using the predicted uncertainty map $\Sigma_{i,j} = \Sigma_i(\mathbf{y}_j)$, following the aleatoric uncertainty formulation of~\citep{wang2025vggt,wang2024dust3r}. The depth regression loss is then formulated as
\begin{align}
    \mathcal{L}_{\text{depth}} &= \frac{1}{HW} \sum_{i,j} \Sigma_{i,j} \, | \! \log \hat{D}_{i,j}  - \log D^\text{gt}_{i,j} | \; - \; \alpha \log \Sigma_{i,j}.
\end{align}
The point map loss is defined following VGGT~\citep{wang2025vggt}: we first represent all quantities in the coordinate system of the first camera, then calculate the mean Euclidean distance $\mathsf{s}$ from the 3D points in the point map $P^{gt}$ to the origin, and use this value as a normalization scale when predicting point map:
\begin{align}
    \mathcal{L}_{\text{pmap}} &= \frac{1}{HW} \sum_{i,j}  \Sigma_{i,j} \, \! \| \hat{P}_{i,j}  - \frac{1}{\mathsf{s}} P^\text{gt}_{i,j} \| \; - \; \alpha \log \Sigma_{i,j}.
\end{align}
We share the same uncertainty map $\Sigma_i$ between the depth and point map losses, since both quantities are derived from the same underlying geometry, and set $\alpha=0.2$ in our experiments.
Since the normal map $\mathbf{N}$ is computed from the depth map and intrinsics via finite differences, it is ill-defined at image boundaries and at pixels lacking valid ground-truth depth. We therefore restrict the normal loss to a validity mask $\mathcal{M}_i \in \{0,1\}^{H\times W}$ marking pixels with well-defined ground-truth normals:
\begin{align}
    \mathcal{L}_{\text{normal}} &= \frac{1}{\sum_{i,j} \mathcal{M}_{i,j}} \sum_{i,j} \mathcal{M}_{i,j} \left(1 - \hat{\mathbf{N}}_{i,j} \cdot \mathbf{N}_{i,j}\right).
\end{align}

\af{\textbf{Training Data.} We train on a mixture of 18 public real and synthetic datasets covering indoor, outdoor, object-centric and driving scenes: ScanNet~\citep{dai2017scannet}, Hypersim~\citep{roberts2021hypersim}, TartanAir and its V2 release~\citep{wang2020tartanair}, Virtual KITTI 2~\citep{cabon2020vkitti2}, BlendedMVG~\citep{yao2020blendedmvs}, MVS-Synth~\citep{huang2018deepmvs}, Taskonomy~\citep{zamir2018taskonomy}, MegaDepth~\citep{li2018megadepth}, CO3D~\citep{reizenstein2021common}, Replica~\citep{straub2019replica}, Aria Synthetic Environments~\citep{avetisyan2024scenescript}, CA-1M~\citep{lazarow2025cubify}, renderings of TRELLIS 3D assets~\citep{xiang2025structured}, UnrealStereo4K~\citep{tosi2021smd}, Spring~\citep{mehl2023spring}, Mapillary Metropolis~\citep{mapillary2021metropolis} and Kubric-4D~\citep{greff2022kubric,van2024generative}. We do not use all the sequences available in these datasets, but only a subset of them. Training views are selected per dataset following DVLT~\citep{burzio2026d}. }

\af{\textbf{Implementation Details.} The image encoder is initialized with the pretrained DINOv3 weights, and the matching features are extracted with the first block of ResNet18~\citep{he2016deep} at $\frac{H}{4} \times \frac{W}{4}$. The transformer has $8$ layers with an embedding dimension of $768$. Following DVLT~\citep{burzio2026d}, each training sample has $2$ to $18$ views, $640$ pixels wide with an aspect ratio drawn per batch from $[0.33, 1]$, and each batch holds as many samples as fit in $20{,}000$ patch tokens per GPU. We use $T=128$ depth hypotheses; the depth range is taken from the ground truth and randomly widened during training, and estimated from the camera baselines with a coarse-to-fine sweep at inference. We train with AdamW (learning rate $10^{-4}$, $10^{-5}$ for the image encoder, weight decay $10^{-4}$, cosine schedule) for $160k$ iterations in bf16 precision on $4$ A100 GPUs over $4$ days.}

\section{Experiments}

\begin{table}[t]
    \renewcommand{\arraystretch}{1.4}
    \begin{center}
    \resizebox{\linewidth}{!}{
    \begin{tabular}{lccccccccccc}
        \toprule
        \multirow{2}{*}[-4pt]{\textbf{Methods}}
        & \multicolumn{2}{c}{\textbf{ETH3D}}
        & \multicolumn{2}{c}{\textbf{KITTI}}
        & \multicolumn{2}{c}{\textbf{DTU}}
        & \multicolumn{2}{c}{\textbf{ScanNet}}
        & \multicolumn{2}{c}{\textbf{Tanks\&Temples}}
        & \multirow{2}{*}[-4pt]{\textbf{Runtime (s)}}
        \\
        \cmidrule(lr){2-3}
        \cmidrule(lr){4-5}
        \cmidrule(lr){6-7}
        \cmidrule(lr){8-9}
        \cmidrule(lr){10-11}
        & AUC.05 & AUC.10
        & AUC.05 & AUC.10
        & AUC.05 & AUC.10
        & AUC.05 & AUC.10
        & AUC.05 & AUC.10
        &
          \\
        \midrule
        VGGT & 0.06 & 0.21 & 0.01 & 0.03 & 0.68 & \third0.84 & 0.34 & 0.59 & 0.02 & 0.08 & 0.499 \\
        Fast3R & 0.02 & 0.10 & 0.00 & 0.02 & 0.16 & 0.31 & 0.15 & 0.38 & 0.01 & 0.06 & \first0.153 \\
        MapAnything & 0.04 & 0.18 & 0.00 & 0.01 & 0.07 & 0.19 & 0.15 & 0.40 & 0.02 & 0.10 & 1.635 \\
        Pi3 & 0.20 & 0.47 & 0.03 & 0.13 & 0.62 & 0.79 & \second0.50 & \second0.72 & 0.03 & 0.13 & \second0.235 \\
        DepthAnything3 & 0.17 & 0.44 & 0.03 & 0.13 & 0.70 & \third0.84 & 0.42 & 0.63 & 0.03 & 0.13 & 0.492 \\
        \af{DVLT} & 0.06 & 0.23 & 0.01 & 0.05 & 0.41 & 0.66 & 0.47 & \third0.69 & 0.02 & 0.09 & 0.277 \\
        VGGT-$\Omega$ & 0.21 & \third0.52 & 0.09 & \third0.33 & \af{0.70} & \third0.84 & 0.46 & 0.68 & 0.03 & 0.12 & 0.296 \\
        \midrule
        MapAnything-w/c & 0.02 & 0.07 & 0.00 & 0.00 & 0.19 & 0.41 & 0.20 & 0.44 & 0.01 & 0.06 & 1.425 \\
        Pi3x-w/c & 0.21 & 0.51 & 0.03 & 0.13 & 0.60 & 0.78 & \first0.52 & \first0.74 & 0.03 & 0.14 & \third0.276 \\
        DepthAnything3-w/c & 0.24 & \second\af{0.55} & 0.08 & 0.31 & 0.72 & \second\af{0.85} & \third0.48 & 0.66 & 0.03 & 0.15 & 0.521 \\
        \midrule
        MVSFormer++ & \third0.30 & 0.46 & \second\af{0.44} & \second0.62 & \third0.80 & 0.82 & 0.16 & 0.25 & \third0.23 & \third0.37 & 2.236 \\
        MVSA & \second0.33 & 0.50 & \third0.42 & \second0.62 & \second0.81 & \third0.84 & 0.35 & 0.53 & \second0.25 & \second0.43 & 2.155 \\
        \af{S2S-MVS} & \first\af{0.37} & \first\af{0.58} & \first\af{0.47} & \first\af{0.67} & \first\af{0.87} & \first\af{0.92} & \af{0.42} & \af{0.62} & \first\af{0.28} & \first\af{0.44} & \af{0.339} \\
        \bottomrule
    \end{tabular}
    }
    \end{center}
    \caption{
    \af{Comparisons with state-of-the-art sequence-to-sequence feed-forward models on reconstruction accuracy using the RMVDB benchmark. The -w/c suffix indicates that the model uses ground truth camera parameters as input; MVSA, MVSFormer++ and S2S-MVS are MVS methods that always require ground truth cameras. The \colorbox{red!25}{best}, \colorbox{orange!25}{second-best} and \colorbox{yellow!35}{third-best} results in each column are highlighted.}}
    \label{tab:comp_recon}
\end{table}

In this section, we evaluate the performances of our model and the state-of-the-art competitors including both sequence-to-sequence and sequence-to-one methods on 3D reconstruction tasks. Following this, we will also present an ablation study to validate the effectiveness of the proposed ray-map embedding and global-cost-volume designs.

\textbf{Baselines.} Our main competitors are the feed-forward sequence-to-sequence methods, including VGGT~\citep{wang2025vggt}, Fast3R~\citep{yang2025fast3r}, MapAnything~\citep{keetha2025mapanything}, Pi3~\citep{wang2025pi}, DepthAnything3~\citep{lin2025depth}, VGGT-$\Omega$~\citep{wang2026vggt} and DVLT~\citep{burzio2026d}. Among these methods, VGGT, Fast3R, VGGT-$\Omega$ and DVLT do not take camera parameters as input, we will use the estimated camera parameters to produce point maps in our evaluation. MapAnything, Pi3 and DepthAnything3 support camera parameters as input, we will additionally evaluate these models with ground truth camera parameters as input, to form direct and fair comparison against our method. All methods take the first-camera frame as the world frame. We also compare against MVSFormer++~\citep{cao2024mvsformer++} and MVSA~\citep{izquierdo2025mvsanywhere}, which are the state-of-the-art sequence-to-one models to the best of our knowledge.

\textbf{Benchmark.} We use the RMVDB benchmark~\citep{schroppel2022benchmark} to evaluate all methods, which consists of five multi-view datasets\af{, i.e.,} KITTI~\citep{geiger2012we}, ScanNet~\citep{dai2017scannet}, ETH3D~\citep{schops2017multi}, DTU~\citep{jensen2014large} and Tanks \& Temples~\citep{knapitsch2017tanks}. \af{Except for ScanNet, whose training split is part of our training data, none of these datasets is used for training.} These datasets cover a diverse set of real-world scenarios, including driving sequences, room scans, building scans, and tabletop objects. We use all 8-frame tuples provided in the RMVDB for evaluation.

\textbf{Metrics.} We use relative per-point errors between predicted and ground truth point maps to compute our main evaluation metric. To account for the different scales in different scenes, we first compute the scale of each predicted or ground truth point maps as the mean distance to the origin, and normalize the point maps accordingly before evaluation. Then the relative per-point error is computed as the point errors divided by the ground-truth depth of that point. We report the robust Area Under the Curve (AUC) statistics of the computed relative per-point errors at finer $0.05$ and coarser $0.10$. 

\begin{figure}[t]
    \centering
    \includegraphics[width=\linewidth]{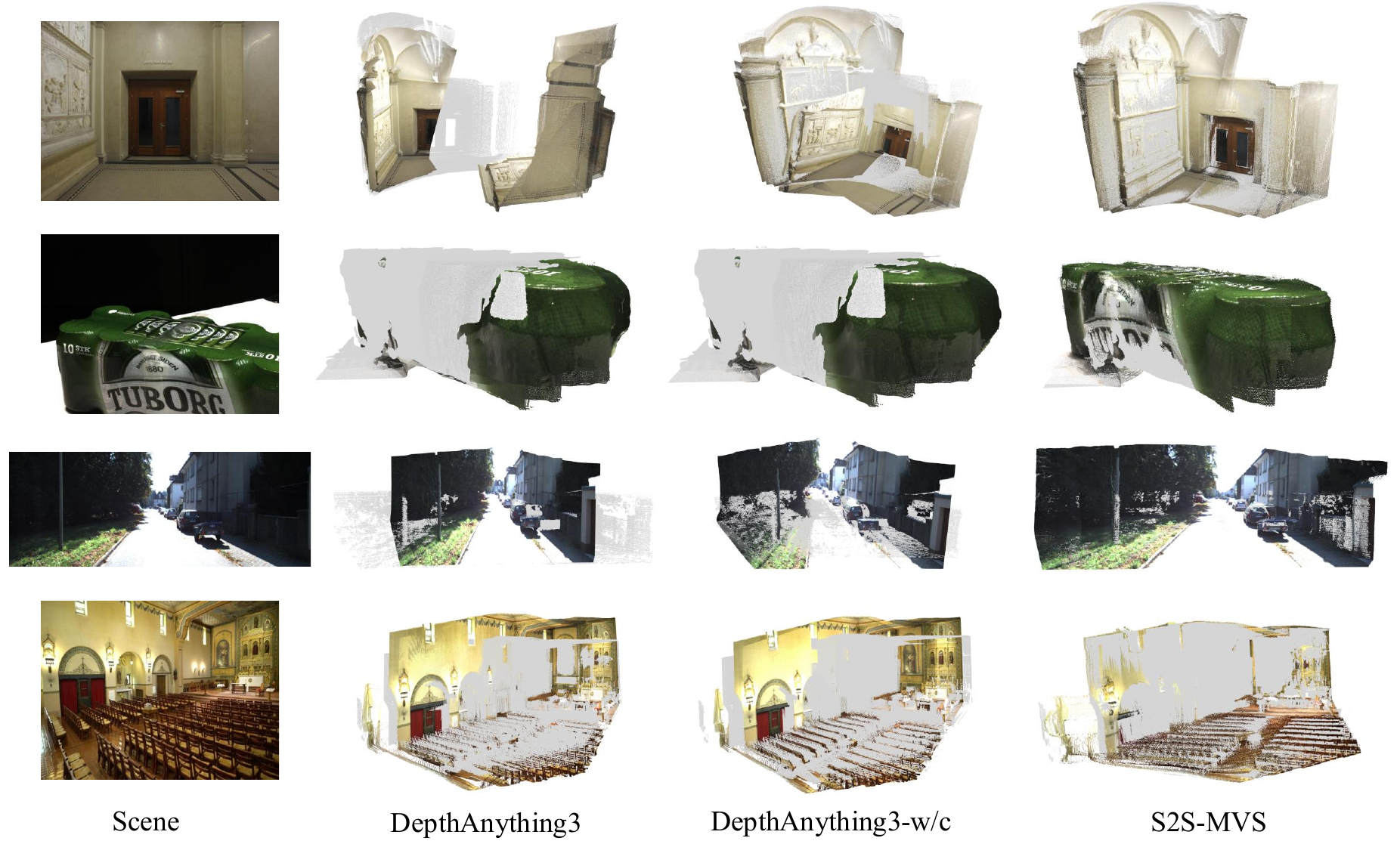}
    \caption{Qualitative point map prediction results of our proposed method compared to DA3. We overlay the predicted point maps and gray-coded ground truth maps to highlight the accuracy of alignment.}
    \label{fig:qualitative}
\end{figure}

\subsection{Comparison with State-of-the-Art Methods}

Table~\ref{tab:comp_recon} compares our S2S-MVS with recent sequence-to-sequence feed-forward reconstruction models and sequence-to-one MVS models on the RMVDB benchmark. Overall, S2S-MVS achieves the best reconstruction accuracy on \af{all datasets except ScanNet}, obtaining the highest AUC.05/AUC.10 scores on ETH3D\af{, KITTI, DTU and Tanks\&Temples} and showing strong generalization to both indoor structured scenes and outdoor driving scenes.

In addition to camera-free feed-forward methods, S2S-MVS also compares favorably to feed-forward and MVS methods that take camera parameters as input, demonstrating the effectiveness of our sequence-to-sequence MVS formulation. \af{On ScanNet, which is part of the training data of most feed-forward models as well as ours, several feed-forward models (e.g., Pi3, VGGT-$\Omega$ and DVLT) obtain higher accuracy, while S2S-MVS still clearly outperforms the other MVS baselines (MVSFormer++ and MVSA).}

We also compare runtime efficiency in Table~\ref{tab:comp_recon}. S2S-MVS runs \af{in the same range as} the sequence-to-sequence feed-forward models and substantially faster than heavier feed-forward pipelines \af{such as MapAnything}. More importantly, compared with the sequence-to-one MVS baselines that achieve the closest reconstruction accuracy to ours, S2S-MVS is considerably faster than both MVSFormer++ and MVSA, while \af{exceeding their accuracy on all five datasets}. This shows that our sequence-to-sequence formulation combines the accuracy of dedicated MVS pipelines with the efficiency of feed-forward reconstruction models.

We also provide qualitative comparisons in Fig.~\ref{fig:qualitative}. In the figure, we compare with DepthAnything3 in both w/ and w/o camera settings, and visualize the predicted point maps overlaid with the gray-coded ground truth point maps. We can see that S2S-MVS produces more accurate reconstructions compared to other methods. The performance is more pronounced on the last scene in the figure from Tanks\&Temples dataset, where the reconstructed geometries of other methods suffer from 3D anisotropic scale distortion, leading to an incorrect aspect ratio of the reconstructed scene. In contrast, our method does not suffer from this issue.

\subsection{Further Analysis}

\begin{figure}[t]
    \centering
    \includegraphics[width=\linewidth]{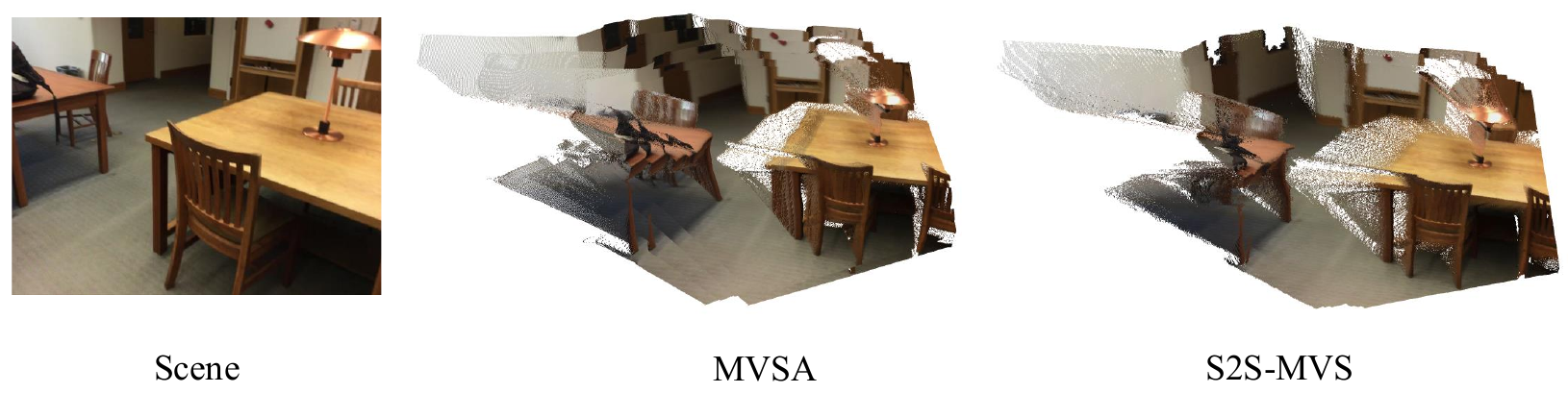}
    \caption{Comparison of point map prediction results between our method and MVSA. The proposed sequence-to-sequence design of S2S-MVS achieves better global alignment of the reconstructed geometry.}
    \label{fig:qualitative2}
\end{figure}

\textbf{Global Alignment.} The benefit of the sequence-to-sequence design of S2S-MVS is that it can achieve better global alignment of the reconstructed geometry across views, as the model can directly learn to reason about the global geometry from the input image sequences. In contrast, sequence-to-one methods like MVSA reconstruct each view independently, which can lead to misalignment issues across views. We provide a qualitative comparison in Fig.~\ref{fig:qualitative2}, where we visualize the predicted point maps from S2S-MVS and MVSA, where our method produces more globally consistent reconstructions across views.

\begin{figure}[t]
    \centering
    \includegraphics[width=\linewidth]{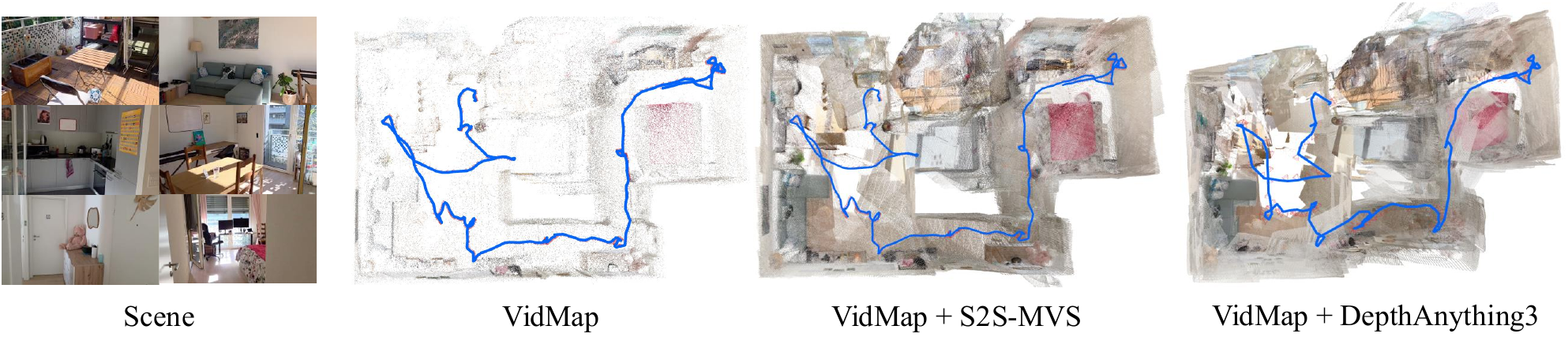}
    \caption{Dense video-based reconstruction results. We use camera parameters estimated by VidMap as input to our S2S-MVS model and DepthAnything3 to predict point maps. Our method produces more accurate reconstructions compared to DepthAnything3, which does not benefit from known camera parameters.}
    \label{fig:sfm}
\end{figure}

\textbf{Combined with SfM.} Our model can be easily combined with Structure-from-Motion (SfM) techniques to accomplish dense reconstruction. In this experiment, we use VidMap~\citep{pataki2026vidmap} as the SfM method to estimate camera parameters from input image sequences extracted from a video. The camera parameters are provided to our S2S-MVS model and also DepthAnything3 to predict point maps. We visualize the  dense point cloud reconstruction results in Fig.~\ref{fig:sfm}. We can see that our method produces more accurate dense reconstructions compared to DepthAnything3, which creates misaligned point clouds and camera offsets.

\subsection{Ablation Study}
We also conduct an ablation study to analyze the impact of different components in our model. The results are summarized in Tab.~\ref{tab:ablation}. We evaluate the variants of our model without the ray-map embedding and without the global cost volume, respectively. We can see that both components contribute significantly to the overall performance of our model. Remove ray map embedding leads to degradation in global alignment, and affect the performance. When training without the global cost volume, the performance drops more significantly, which shows the importance of the global cost volume design in providing strong regularization for the model.

\begin{table}[h]
    \begin{center}
    \resizebox{\linewidth}{!}{
    \begin{tabular}{lccccccccccc}
        \toprule
        \multirow{2}{*}[-4pt]{\textbf{Methods}} 
        & \multicolumn{2}{c}{\textbf{ ETH3D }} 
        & \multicolumn{2}{c}{\textbf{ KITTI }} 
        & \multicolumn{2}{c}{\textbf{ DTU }} 
        & \multicolumn{2}{c}{\textbf{ ScanNet }} 
        & \multicolumn{2}{c}{\textbf{ Tanks\&Temples }} 
        \\
        \cmidrule(lr){2-3} 
        \cmidrule(lr){4-5} 
        \cmidrule(lr){6-7} 
        \cmidrule(lr){8-9} 
        \cmidrule(lr){10-11} 
        & AUC@.05 & AUC@.10
        & AUC@.05 & AUC@.10
        & AUC@.05 & AUC@.10
        & AUC@.05 & AUC@.10
        & AUC@.05 & AUC@.10
          \\
        \midrule
        w/o Ray Map Embedding & 0.35 & 0.55 & 0.44 & 0.62 & 0.80 & 0.88 & 0.35 & 0.50 & 0.24 & 0.40 \\
        w/o Global Cost Volume & 0.25 & 0.46 & 0.33 & 0.53 & 0.61 & 0.72 & 0.30 & 0.47 & 0.20 & 0.38 \\ 
        Full Model & \af{0.37} & \af{0.58} & \af{0.47} & \af{0.67} & \af{0.87} & \af{0.92} & \af{0.42} & \af{0.62} & \af{0.28} & \af{0.44} \\ 

        \bottomrule
    \end{tabular}
    }
    \end{center}
    \caption{Ablation study on the effectiveness of the proposed ray map embedding and global cost volume designs.}
    \label{tab:ablation}
\end{table}

\section{Discussion and Conclusion}
In this paper, we propose S2S-MVS, a novel sequence-to-sequence feed-forward model for multi-view stereo reconstruction. Our model is designed to directly predict \af{depth maps for each view from input image sequences, from which point maps are obtained with the ray maps,} assuming camera parameters to be known. We introduce a ray-map embedding design to encode the camera information of the input views, and a global cost volume design to provide strong regularization for geometry prediction. We evaluate our model on the RMVDB benchmark, and show that it achieves superior reconstruction accuracy compared to state-of-the-art sequence-to-sequence feed-forward models. However, compared to other feed-forward models, our model is still limited by the scale of training data and thus does not \af{generalize} well in certain scenarios. Nonetheless, we believe that our S2S-MVS model provides a promising direction for efficient and accurate multi-view stereo reconstruction, and can be further improved by exploring more advanced architectures and training strategies.

\begin{ack}
This work was funded in part by the Swiss National Science Foundation.
\end{ack}

\bibliographystyle{plainnat}
\bibliography{new}

\end{document}